\documentclass[twoside]{article}

\usepackage[margin=3cm]{geometry}
\usepackage{amsmath}
\usepackage{amssymb}
\usepackage[ansinew]{inputenc}
\usepackage{bm}
\usepackage{float}
\usepackage{url}
\usepackage{algorithm}
\usepackage{algorithmic}
\usepackage{graphicx}
\usepackage{subfigure}
\usepackage[small]{caption}
\usepackage{url}
\usepackage{verbatim}
\usepackage[authoryear]{natbib}
\usepackage{times}

\newenvironment{matlab}{\comment}{\endcomment}

\begin{matlab}

addpath(genpath('anc'))
clear all
close all
rand('seed', 1e5);
randn('seed', 1e5);
fontName = 'times';
fontSize = 26;

load DellaGattadata.mat 
N = size(exprs_tp53_RMA,2);
tTrue = [timepoints; truncTimepoints]; 
typDiff = 20; 
kern.inverseWidth = 1/(typDiff*typDiff); 
noise = 1e-4;
n2 = (ones(size(tTrue))*(tTrue.^2)')' + ones(size(tTrue))*(tTrue.^2)' - 2.*tTrue*tTrue';
wi2 = (.5 .* kern.inverseWidth);
K_comb = exp(-n2*wi2) + eye(20)*noise;
idx_top = 1:length(exprs_tp53_RMA);
Y = [exprs_tp53_RMA(:, idx_top); exprs_null_RMA(:, idx_top)]; 
Y = Y - repmat(mean(Y,2),1, size(Y,2)); 
Cy = Y*Y'/size(Y,2); 
[S D] = eig(Cy, K_comb); 
[D perm] = sort(diag(D),'descend');
Y = [exprs_tp53_RMA; exprs_null_RMA]; 
Y = Y - repmat(mean(Y,2),1, N); 
Xproj = S(:,perm(D>1))'*Y; 
dists = sum(Xproj.^2,1); 

figure(1), clf, colormap gray
imagesc(K_comb), colorbar
set(gca, 'fontname', fontName, 'fontsize', fontSize), daspect('manual')
print -depsc RCAonDGATTAdata_combRBFkernel_noise1e-4.eps
system('epstopdf RCAonDGATTAdata_combRBFkernel_noise1e-4.eps');

figure(2), clf
compareROC(dists, DGatta_labels_byTSNItop100, BATSrankingFilepath); 
set(gca, 'fontname', fontName, 'fontsize', fontSize), daspect('manual')
print -depsc RCAvsBATS_onTP63_noise1e-4.eps
system('epstopdf RCAvsBATS_onTP63_noise1e-4.eps');

dataSetName = 'silhouette';
converged = false;
tracesC = [];
k = 1;  t = 1; i = 1;
limit = 1e-3;
[X, Y, XTest, YTest] = mapLoadData(dataSetName); 
[n, sx] = size(X);
sy = size(Y, 2);
noise = .03;
X = X + randn(size(X))*noise; 
XTest = XTest + randn(size(XTest))*noise;
Y = Y + randn(size(Y))*noise;
YTest = YTest + randn(size(YTest))*noise;
Sy = cov(Y, 1); Sx = cov(X, 1); S = cov([X Y], 1);  
Sy = Sy + abs(mean(diag(Sy)))*1e-6*eye(sy); S(sx+1:end,sx+1:end) = Sy;  
Sx = Sx + abs(mean(diag(Sx)))*1e-6*eye(sx); S(1:sx,1:sx) = Sx;
for alpha = .05:.05:1;  
    noisey = alpha*trace(Sy)/sy;    noisex = alpha*trace(Sx)/sx;
    Ny = noisey*eye(sy); Nx = noisex*eye(sx); Noise = blkdiag(Nx,Ny);  
    W1W1 = zeros(sy);   W4W4 = zeros(sx);   
    W2W2 = zeros(sy);   W3W3 = zeros(sx);   
    W2 = zeros(sy);     W3 = zeros(sx,sy);
    dz1 = 0;            dz2 = size(W2W2,2);    dz3 = 0;
    disp ' '; diagnostic;
    oldLML = LML_RCA;
    while ~converged
        [W1 Ly] = eig(Sy, W2W2 + Ny);           [Ly permy] = sort(diag(Ly), 'descend');
        [W4 Lx] = eig(Sx, W3W3 + Nx);           [Lx permx] = sort(diag(Lx), 'descend');
        W1 = (W2W2 + Ny) * W1(:,permy(Ly>1)) * sqrt(diag(Ly(Ly>1)-1));     
        W4 = (W3W3 + Nx) * W4(:,permx(Lx>1)) * sqrt(diag(Lx(Lx>1)-1));
        W1W1 = W1*W1';                          W4W4 = W4*W4';
        dz1 = size(W1,2);                       dz3 = size(W4,2);
        [W r] = eig(S, blkdiag(W4W4, W1W1) + Noise);    [r perm] = sort(diag(r), 'descend');
        W = (blkdiag(W4W4, W1W1) + Noise) * W(:,perm(r>1)) * sqrt(diag(r(r>1)-1)); 
        W2 = W(sx+1:end,:);     W3 = W(1:sx,:);
        W2W2 = W2*W2';          W3W3 = W3*W3';
        dz2 = size(W,2);
        disp('## RCA for W2, W3 jointly'), diagnostic;
        k = k + 1;
        converged = (LML_RCA - oldLML) < limit;
        oldLML = LML_RCA;
    end
    converged = false;
    stats.alpha(i) = alpha;
    stats.RMSerror(i) = RMSerror_RCA;
    stats.dz1(i) = dz1;
    stats.dz2(i) = dz2;
    stats.dz3(i) = dz3;
    i = i + 1;
end

for d = 1:57
    [MLE E var CCA] = pcca(X,Y,d);
    YPred_pCCA = XTest*pdinv(Sx)*MLE.Wzx * MLE.Wzy' + repmat(mean(Y,1), size(XTest,1), 1);
    Ysqdiff = (YTest - YPred_pCCA).^2;
    RMSerror_pCCA(d) = sqrt(sum(Ysqdiff(:))/numel(Ysqdiff)); 
end

figure(1), clf
subplot(211)
plot(stats.alpha, stats.RMSerror, 'go-', 'Color', [0 .5 0], 'LineWidth', 3, 'MarkerFaceColor', [0 .5 0], 'MarkerSize', 5)
legend 'Iterative RCA'
set(gca, 'fontname', fontName, 'fontsize', fontSize)
xlabel('\alpha', 'fontName', fontName, 'fontSize', fontSize)
ylabel('RMS error', 'fontName', fontName, 'fontSize', fontSize), 
ylim([3 3.5])
subplot(212)
plot(RMSerror_pCCA, 'ko-', 'LineWidth', 3, 'MarkerFaceColor', [0 0 0], 'MarkerSize', 5)
legend('PCCA','location','southeast')
xlabel('d', 'fontName', fontName, 'fontSize', fontSize)
ylabel('RMS error', 'fontName', fontName, 'fontSize', fontSize)
ylim([3 3.5])
set(gca, 'fontname', fontName, 'fontsize', fontSize)
print -depsc RCAvsPCCA_noise3e-2.eps
system('epstopdf RCAvsPCCA_noise3e-2.eps')

figure(2), clf, hold on
plot(stats.alpha, stats.dz1, 'ro:', 'Color', [.75 0 0], 'LineWidth', 4, 'MarkerSize', 4)
plot(stats.alpha, stats.dz2, 'go-', 'Color', [0 .5 0], 'LineWidth', 4, 'MarkerSize', 4)
plot(stats.alpha, stats.dz3, 'bo--', 'Color', [0 0 .75], 'LineWidth', 4, 'MarkerSize', 4)
daspect([1 100 1])
xlabel('\alpha', 'fontName', fontName, 'fontSize', fontSize)
ylabel('Retained dimensions', 'fontName', fontName, 'fontSize', fontSize)
legend 'X1' 'Z' 'X2'
set(gca, 'fontname', fontName, 'fontsize', fontSize)
print -depsc RCAretdim_noise3e-2.eps
system('epstopdf RCAretdim_noise3e-2.eps')

\end{matlab}

\title{Residual Component Analysis}

\author{
Alfredo A. Kalaitzis, Neil D. Lawrence\\
Sheffield Institute for Translational Neuroscience\\
and\\ Department of Computer Science\\
University of Sheffield\\
\texttt{\{A.Kalaitzis,N.Lawrence\}@sheffield.ac.uk}
}

\begin{document}

\maketitle

\begin{abstract}
Probabilistic principal component analysis (PPCA) seeks a low
dimensional representation of a data set in the presence of
independent spherical Gaussian noise, $\mathbf{\Sigma} =
\sigma^2\mathbf{I}$. The maximum likelihood solution for the model
is an eigenvalue problem on the sample covariance matrix. In this
paper we consider the situation where the data variance is already
partially explained by other factors, e.g. covariates of interest,
or temporal correlations leaving some residual variance. We
decompose the residual variance into its components through a
generalized eigenvalue problem, which we call residual component
analysis (RCA). We show that canonical covariates analysis (CCA) is
a special case of our algorithm and explore a range of new
algorithms that arise from the framework. We illustrate the ideas on
a gene expression time series data set and the recovery of human
pose from silhouette.
\end{abstract}

\section{Introduction} \label{sec:intro}

Probabilistic principal component analysis (PPCA) decomposes the covariance of a
data point, $\mathbf{y}$, into a low rank term and a diagonal noise term. The
underlying probabilistic model assumes that each datum is Gaussian distributed,
\[
\mathbf{y} \sim \mathcal{N}(\mathbf{0}, \mathbf{W}\mathbf{W}^\top + \sigma^2
\mathbf{I}),
\]
where we assume the data is centred such that its mean is zero and
$\mathbf{W}\in \Re^{d\times q}$ imposes a reduced rank structure on the
covariance ($q<d-1$). The log likelihood of the centered data set with $n$ data
points, $\mathbf{Y}\in \Re^{n\times d}$,
\[
  \log p(\mathbf{Y}) = \sum_{i=1}^n \log
  \mathcal{N}(\mathbf{y}_{i, :}|\mathbf{0}, \mathbf{W}\mathbf{W}^\top
  + \sigma^2 \mathbf{I}),
\]
can be maximized \citep{Tipping:ppca96} with the result that
$\mathbf{W}=\mathbf{U}_q\mathbf{L}_q\mathbf{R}^\top$, where
$\mathbf{U}_q$ are the $q$ principal eigenvectors of the sample
covariance, $\mathbf{S}=n^{-1}\mathbf{Y}^\top\mathbf{Y}$, $\mathbf{L}_q$
is a diagonal matrix with elements
$\ell_{i,i}=\sqrt{\lambda_i-\sigma^2_n}$, where $\lambda_i$ is the
$i$th eigenvalue of the sample covariance, $\mathbf{R}$ is an
arbitrary rotation matrix, and $\sigma^2_n$ the noise variance. As a result the
matrix $\mathbf{W}$ spans the principal subspace of the data and the model is
known as principal components analysis.  Underlying this model is an assumption
that the data set can be represented by
\[
\mathbf{Y} = \mathbf{X}\mathbf{W}^\top + \mathbf{E}
\]
where $\mathbf{X} \in \Re^{n\times q}$ is the matrix of $q$-dimensional
latent variables and $\mathbf{E}$ is a matrix of noise variables, each
element being independently sampled from a zero mean Gaussian with
variance $\sigma^2$. The marginal likelihood above is obtained by
placing an isotropic prior independently on the elements of
$\mathbf{X}$, $x_{i,j}\sim\mathcal{N}(0, 1)$.

Lawrence \citep{Lawrence:pnpca06} showed that the PCA solution is also obtained
for log likelihoods of the form
\[
\log p(\mathbf{Y}) = \sum_{j=1}^d \log \mathcal{N}(\mathbf{y}_{:, j}|\mathbf{0},
\mathbf{X}\mathbf{X}^\top + \sigma^2\mathbf{I})
\]
which is recovered when we marginalize $\mathbf{W}$ with an isotropic
prior instead of $\mathbf{X}$. This is a dual\footnote{As opposed to the
typical primal form. Refers to the duality between the data-space (row-space)
and the coordinate-space (column-space) of a design matrix, with data-samples as
its rows.} form of probabilistic PCA which could also be called probabilistic
principal coordinate analysis as the maximum likelihood solution solves for the
latent coordinates, $\mathbf{X}=\mathbf{U}^\prime_q\mathbf{L}\mathbf{R}^\top$,
instead of the principal subspace. Here $\mathbf{U}^\prime_q$ are the first $q$
principal eigenvectors of the inner product matrix $\mathbf{Y}\mathbf{Y}^\top$
with $\mathbf{L}$ defined as before. Note in this case that the Gaussian density
is independent across \emph{data features} rather than \emph{data points}. So
the correlation is expressed between data points. The underlying model is in
fact an product of independent Gaussian processes \citep{Rasmussen:book06} with
linear covariance functions.

Both of these scenarios involve maximizing log likelihoods of a
similar structure, namely the covariance of the Gaussians is given by
a low rank term plus a spherical term, $\mathbf{X}\mathbf{X}^\top + \sigma^2
\mathbf{I}$ (dual scenario). In this paper we consider an alternative
form where the covariance is given by $\mathbf{X}\mathbf{X}^\top +
\mathbf{\Sigma}$,
where $\mathbf{\Sigma}$ is \emph{a general positive definite matrix}. Our
motivation is that our data has already been partly explained by the
covariance matrix $\mathbf{\Sigma}$ and we wish to study the components of the
residuals. Our ideas can be applied in both the primal and dual
representations: the form to be used depends on what information we
wish to include in $\mathbf{\Sigma}$.

As a motivating example consider a linear additive model (Figure \ref{fig:rca}),
\begin{equation} \label{eq:linearmodel}
\mathbf{Y} = \mathbf{X}\mathbf{W}^\top + \mathbf{Z}\mathbf{V}^\top + \mathbf{E},
\end{equation}
where $\mathbf{Z}$ is a matrix of known covariates that are assumed to have some
predictive power for $\mathbf{Y}$ and $\mathbf{X}$ is a matrix of unknown
confounders (as in standard PPCA). Also consider that $\mathbf{Y}$
could be a set of $n$ patients' gene expression measurements ($d$ genes),
$\mathbf{Z}$ could be the genotype of each patient and $\mathbf{X}$ could be
unobserved environmental confounders (see \citep{Fusi:}). We can marginalize out
$\mathbf{V}$ with a Gaussian prior, $v_{i, j} \sim \mathcal{N}(0, \alpha)$ as
well as $\mathbf{W}$ with a prior $w_{i, j} \sim \mathcal{N}(0,1)$
\footnote{There is no loss of generalisation by using a standard Gaussian prior
here, since the functional form of $\mathbf{Y}$'s distribution remains unchanged
for a general Gaussian prior.} and recover
\begin{equation*}  \label{eq:marginalY}
\log p(\mathbf{Y}) = \sum_{j=1}^d \log \mathcal{N}(\mathbf{y}_{:, j}|\mathbf{0},
\mathbf{X}\mathbf{X}^\top + \mathbf{\Sigma}),
\end{equation*}
where for this example
$\mathbf{\Sigma}=\alpha \mathbf{Z}\mathbf{Z}^\top + \sigma^2\mathbf{I}$.

Given $\mathbf{\Sigma}$, can we solve for $\mathbf{X}$? As we will
show, the solution for $\mathbf{X}$ is given by a generalized
eigenvalue problem. By using different forms for $\mathbf{\Sigma}$ we can
formulate different models. For example, for a particular choice of
$\mathbf{\Sigma}$ we recover canonical correlates analysis (CCA, see Section
\ref{subsec:CCAequiv}). In the next section we show how the low rank term can be
optimized for general $\mathbf{\Sigma}$. The only constraints that we place on
$\mathbf{\Sigma}$ are that it should be positive definite and invertible.

\section{Optimizing the Likelihood} \label{sec:optimising}

The log likelihood for the RCA model is given by
\begin{equation} \label{eq:logmarginalY}
  L(\mathbf{X}, \mathbf{\Sigma}) = -\frac{d}{2}\ln \left |{\mathbf{K}} \right | -
  \frac{1}{2}\mathrm{tr}({\mathbf{Y}\mathbf{Y}^\top \mathbf{K}^{-1}})
  - \frac{nd}{2}\ln(2\pi),
\end{equation} 
where we have defined $\mathbf{K} \equiv \mathbf{X}\mathbf{X}^\top +
\mathbf{\Sigma}$. We now take the eigendecomposition of $\mathbf{\Sigma}$,
\begin{equation} \label{equ:sigmadecomp}
  \mathbf{\Sigma} = \mathbf{U}\mathbf{\Lambda}\mathbf{U}^\top,
\end{equation}
where $\mathbf{U}^\top \mathbf{U} = \mathbf{I}$ and $\mathbf{\Lambda}$ is a
diagonal matrix. We now project the covariance onto this eigenbasis scaling with
the eigenvalues,
\begin{equation} \label{eq:projectedK}
\hat{\mathbf{K}} =
\mathbf{\Lambda}^{-\frac{1}{2}} \mathbf{U}^\top
    \mathbf{K}\mathbf{U\Lambda}^{-\frac{1}{2}} = 
\mathbf{\Lambda}^{-\frac{1}{2}}\mathbf{U}^\top \mathbf{X}\mathbf{X}^\top
\mathbf{U\Lambda}^{-\frac{1}{2}} + \mathbf{I}.
\end{equation} 
This allows us to define
\begin{equation} \label{eq:kappahat}
  \hat{\mathbf{K}} \equiv \hat{\mathbf{X}}\hat{\mathbf{X}}^\top + \mathbf{I},
  \quad \textrm{where} \quad
  \hat{\mathbf{X}} \equiv
    \mathbf{\Lambda}^{-\frac{1}{2}} \mathbf{U}^\top\mathbf{X},
\end{equation}
and also implies the inverse 
\begin{equation} \label{eq:kappahatinv}
  \hat{\mathbf{K}}^{-1}=\mathbf{\Lambda}^{\frac{1}{2}}\mathbf{U}^\top
  \mathbf{K}^{-1} \mathbf{U\Lambda}^{\frac{1}{2}}.
\end{equation}
Now we note from eq. (\ref{eq:projectedK}) that 
\begin{equation*} \label{eq:det}
  \left|{\mathbf{K}}\right| =
  \hat{\left|\mathbf{K}\right|} \left|\mathbf{\Lambda}\right|,
\end{equation*}
and from eq. (\ref{eq:kappahatinv}) that
\begin{equation*} \label{eq:trace}
  \mathrm{tr}(\mathbf{Y}\mathbf{Y}^\top\mathbf{K}^{-1}) =
  \mathrm{tr}(\mathbf{\Lambda}^{-\frac{1}{2}}\mathbf{U}^\top
    \mathbf{Y}\mathbf{Y}^\top\mathbf{U}
    \mathbf{\Lambda}^{-\frac{1}{2}}\hat{\mathbf{K}}^{-1}),
\end{equation*}
leading us to define
$\hat{\mathbf{Y}} \equiv
\mathbf{\Lambda}^{-\frac{1}{2}}\mathbf{U}^\top\mathbf{Y}$ so that we can rewrite
the entire likelihood from eq. (\ref{eq:logmarginalY}) as
\begin{equation*} \label{eq:translmlY}
  L(\mathbf{X}, \mathbf{\Lambda}, \mathbf{U}) =
  -\sum_{i=1}^n d \ln \lambda_i
  -\frac{d}{2}\ln \hat{\left|\mathbf{K}\right|} -
  \frac{1}{2}\mathrm{tr} (\mathbf{\hat{Y}} \mathbf{\hat{Y}^\top}
  \hat{\mathbf{K}}^{-1}) -\frac{nd}{2}\ln(2\pi).
\end{equation*}

We know how to maximize this new likelihood with respect to $\hat{\mathbf{X}}$.
Following a similar route to the maximum likelihood solution proof in
\citep{Tipping:ppca96}, we take the gradient of the likelihood with respect to
$\mathbf{\hat{X}}$
\begin{align} \label{equ:derivative2}
  \begin{split}
  &\frac{\partial L}{\partial \hat{\mathbf{X}}} = \hat{\mathbf{K}}^{-1}
    \hat{\mathbf{Y}}\hat{\mathbf{Y}}^\top\hat{\mathbf{K}}^{-1}\hat{\mathbf{X}}
    - \hat{\mathbf{K}}^{-1}\hat{\mathbf{X}} = \mathbf{0}, \quad
  \textrm{to give the stationary point} \quad
  \hat{\mathbf{Y}}\hat{\mathbf{Y}}^\top\hat{\mathbf{K}}^{-1}\hat{\mathbf{X}}
  = \hat{\mathbf{X}}.
  \end{split}
\end{align}
By singular value decomposition on $\hat{\mathbf{X}}$, we get
\begin{equation} \label{equ:latentdecomp}
  \hat{\mathbf{X}} = \hat{\mathbf{V}}\mathbf{L}\mathbf{R}^\top,
\end{equation}
then by substituting $\mathbf{\hat{X}}$ in eq. (\ref{equ:derivative2}) and eq.
(\ref{eq:kappahat})
\begin{equation*}
  \begin{split}
  \hat{\mathbf{V}}\mathbf{L}\mathbf{R}^\top &= 
  \hat{\mathbf{Y}}\hat{\mathbf{Y}}^\top
  (\hat{\mathbf{V}}\mathbf{L}^{2}\hat{\mathbf{V}}^\top + \mathbf{I})^{-1}
  \hat{\mathbf{V}}\mathbf{L}\mathbf{R}^\top \\
  \hat{\mathbf{V}}(\mathbf{L}^{2} + \mathbf{I}) &=
  \hat{\mathbf{Y}}\hat{\mathbf{Y}}^\top\hat{\mathbf{V}},
  \end{split}
\end{equation*}
where we make use of the \emph{Woodbury matrix indentity}. Now we see that
maximisation relies on a regular eigenvalue problem of the form
\begin{equation}\label{equ:eigenproblem}
\begin{split}
  \hat{\mathbf{Y}}\hat{\mathbf{Y}}^\top\hat{\mathbf{V}} &= 
  \hat{\mathbf{V}}\mathbf{D}, \quad \textrm{where} \\
  \mathbf{D} &\equiv (\mathbf{L}^{2} + \mathbf{I}).
\end{split}
\end{equation}

We now express this eigenvalue problem in terms of $\mathbf{Y}\mathbf{Y}^\top$.
Substituting $\mathbf{X} = \mathbf{U\Lambda}^{\frac{1}{2}}\hat{\mathbf{X}}$ and
by eq. (\ref{equ:latentdecomp}), we get a decomposition of $\mathbf{X}$ with
the same singular values as $\mathbf{\hat{X}}$
\begin{equation} \label{eq:svdrelations}
  \mathbf{X} =
  \mathbf{U\Lambda}^{\frac{1}{2}}\hat{\mathbf{V}}\mathbf{L}\mathbf{R}^\top =
  \mathbf{T}\mathbf{L}\mathbf{R}^\top,
\end{equation}
where we have defined
$\hat{\mathbf{V}} \equiv \mathbf{\Lambda}^{-\frac{1}{2}}\mathbf{U^\top T}$.
Substituting for $\mathbf{\hat{Y}}$ and $\mathbf{\hat{V}}$ in the
eigenvalue problem from eq.  (\ref{equ:eigenproblem}), recovers the
eigenvalue problem in the original dual-space
\begin{equation*}
\begin{split}
  \mathbf{\Lambda}^{-\frac{1}{2}}\mathbf{U}^\top
  \mathbf{Y}\mathbf{Y}^\top\mathbf{U\Lambda}^{-1} \mathbf{U}^\top \mathbf{T} &=
\mathbf{\Lambda}^{-\frac{1}{2}}\mathbf{U}^\top \mathbf{T}\mathbf{D} \\
\mathbf{Y}\mathbf{Y}^\top\mathbf{\Sigma}^{-1}\mathbf{T} &= \mathbf{T}\mathbf{D},
\end{split}
\end{equation*}
which follows from the inverse of eq. (\ref{equ:sigmadecomp}).

So far, $\mathbf{T}$ is solved via a non-symmetric eigenvalue problem. Assuming
that $\mathbf{\Sigma}$ is positive-definite (i.e. invertible), we define
$\mathbf{S} \equiv \mathbf{\Sigma}^{-1}\mathbf{T}$ and get
\begin{equation*} \label{equ:rcaeigenproblem}
  \mathbf{Y}\mathbf{Y}^\top\mathbf{S} = \mathbf{\Sigma}\mathbf{S}\mathbf{D}
\end{equation*}
which is in the desired form of a \emph{generalized eigenvalue problem}. Now
we can recover $\mathbf{X}$, up to rotation ($\mathbf{R=I}$), via the first $q$
generalised eigenvectors of $\mathbf{Y}\mathbf{Y}^\top$ and eq.
(\ref{eq:svdrelations})
\begin{equation*} \label{eq:recoveredX}
  \mathbf{X} = \mathbf{T}\mathbf{L} = \mathbf{\Sigma}\mathbf{S}\mathbf{L} =
    \mathbf{\mathbf{\Sigma}}\mathbf{S}\mathbf{(D - I)}^{\frac{1}{2}}.
\end{equation*}

Due to the algebraic symmetry between our dual and primal formulations of the
log-marginal likelihood, we can easily extend our derivations to the primal
representation. For example, in the linear model in eq. (\ref{eq:linearmodel})),
the maximum likelihood solution of $\mathbf{W}$ is computed through
\begin{equation} \label{eq:rcaeigenproblem_primal}
  \mathbf{Y^\top}\mathbf{Y}\mathbf{S} = \mathbf{\Sigma}\mathbf{S}\mathbf{D},
  \quad \textrm{where} \quad
  \mathbf{\Sigma} = \alpha \mathbf{V}\mathbf{V}^\top + \sigma^2\mathbf{I}
  \quad \textrm{and} \quad
  \mathbf{W} = \mathbf{\Sigma}\mathbf{S}\mathbf{(D - I)}^{\frac{1}{2}}.
\end{equation}

\subsection{Equivalence to CCA} \label{subsec:CCAequiv} Canonical
covariates analysis is solved through a generalized eigenvalue problem
\citep{de2005eigenproblems,Bach:kica02}.
\begin{equation*} \label{eq:ccaform1}
  \begin{pmatrix} \mathbf{0} & \mathbf{C}_{12}\\ 
  \mathbf{C}_{12}^\top & \mathbf{0} \end{pmatrix}
  \begin{pmatrix} \mathbf{S}_1\\ 
  \mathbf{S}_2 \end{pmatrix}
  = \begin{pmatrix} \mathbf{C_{11}} & \mathbf{0} \\
  \mathbf{0}& \mathbf{C}_{22} \end{pmatrix}
  \begin{pmatrix} \mathbf{S}_1\\ 
  \mathbf{S}_2 \end{pmatrix}
  \mathbf{\Lambda},
\end{equation*}
which can be rewritten as
\begin{equation*} \label{eq:ccaform2}
  \begin{pmatrix}  \mathbf{C}_{11} & \mathbf{C}_{12}\\
  \mathbf{C}_{12}^\top & \mathbf{C}_{22} \end{pmatrix}
  \begin{pmatrix} \mathbf{S}_1\\ 
  \mathbf{S}_2 \end{pmatrix}
  = \begin{pmatrix} \mathbf{C}_{11} & \mathbf{0}\\ 
  \mathbf{0}& \mathbf{C}_{22} \end{pmatrix}
  \begin{pmatrix} \mathbf{S}_1\\ 
  \mathbf{S}_2 \end{pmatrix}
  (\mathbf{\Lambda + I}).
\end{equation*}
A few notes on CCA: The left-most block matrix is the sample covariance matrix
of the joint (augmented) design matrix
$
  \mathbf{C} = n^{-1}(\mathbf{Y}_1,\mathbf{Y}_2)^\top(\mathbf{Y}_1,\mathbf{Y}_2)
  \quad \textrm{and} \quad
  \mathbf{C}_{11} = n^{-1}\mathbf{Y}_1^\top\mathbf{Y}_1,  \quad
  \mathbf{C}_{22} = n^{-1}\mathbf{Y}_2^\top\mathbf{Y}_2, \quad
  \mathbf{C}_{12} = n^{-1}\mathbf{Y}_1^\top\mathbf{Y}_2
$
are the individual sample covariances and cross-covariance of
$\mathbf{Y}_1, \mathbf{Y}_2$. The diagonal matrix of generalised eigenvalues,
$\mathbf{\Lambda}$, contains the \emph{canonical correlations}. The generalised
eigenvectors, made up of direction-pairs $\mathbf{S}_1,\mathbf{S}_2$, are the
\emph{canonical-directions} or \emph{coefficients} in data-spaces
$\mathcal{Y}_1,\mathcal{Y}_2$ respectively. They maximise the correlation
between a projection $\mathbf{Y}_1\mathbf{S}_1$ of features of $\mathbf{Y}_1$
and a projection $\mathbf{Y}_2\mathbf{S}_2$ of features of $\mathbf{Y}_2$,
\[
\mathbf{S}_1^\top\mathbf{C}_{12}\mathbf{S}_2 = \mathbf{P},
\quad \textrm{such that} \quad
\mathbf{S}_1^\top\mathbf{C}_{11}\mathbf{S}_1 =
\mathbf{S}_2^\top\mathbf{C}_{22}\mathbf{S}_2 = \mathbf{I},
\]
where $\mathbf{P}$ is a rectangular matrix with the canonical correlations on
its diagonal. These projections
are known as the \emph{canonical variates}.

\begin{figure}[tb]
\centering
\subfigure[]{ \label{fig:rca}
  \includegraphics[width=0.22\linewidth]{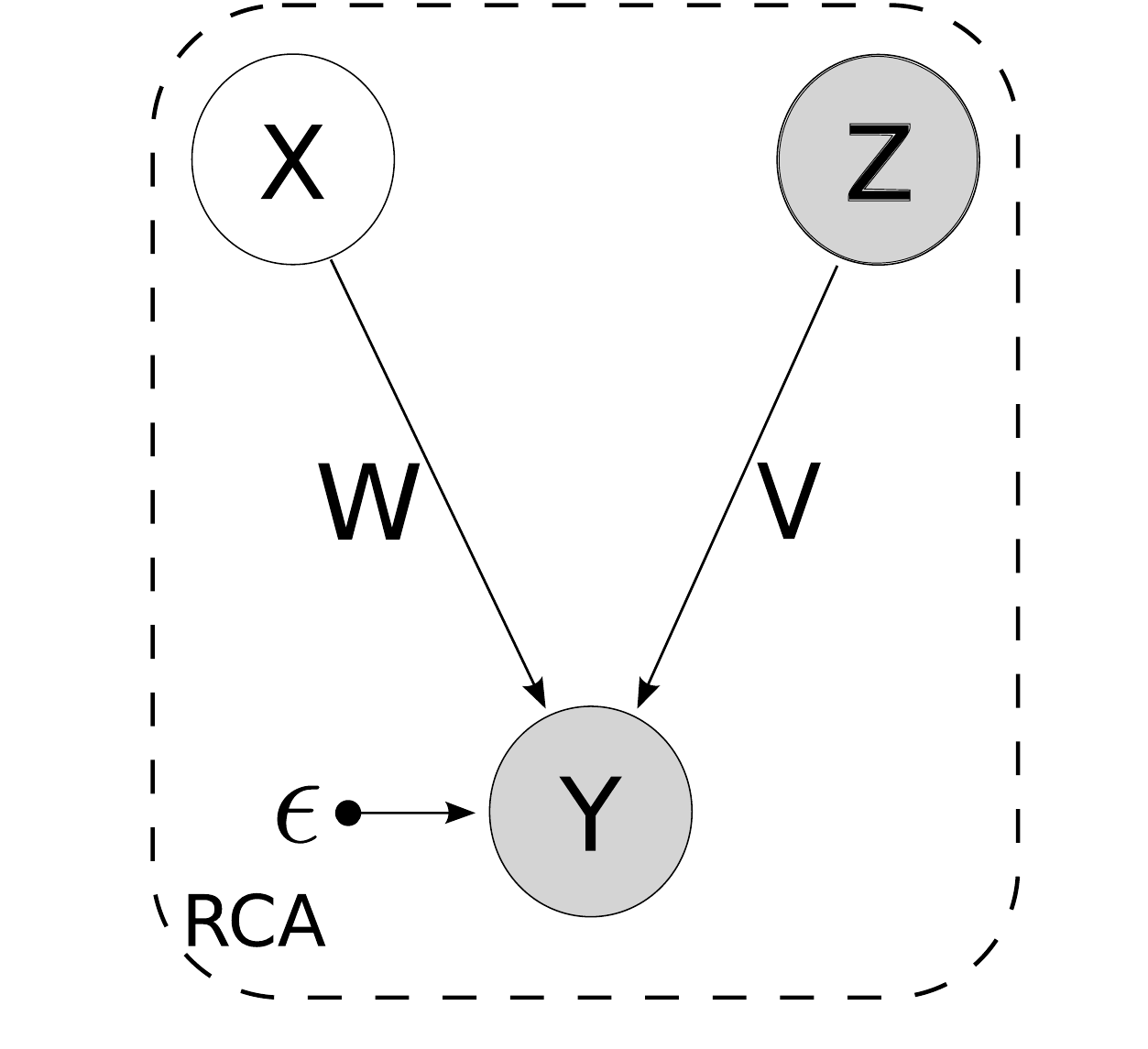}
}
\subfigure[]{ \label{fig:cca}
  \includegraphics[width=0.29\linewidth]{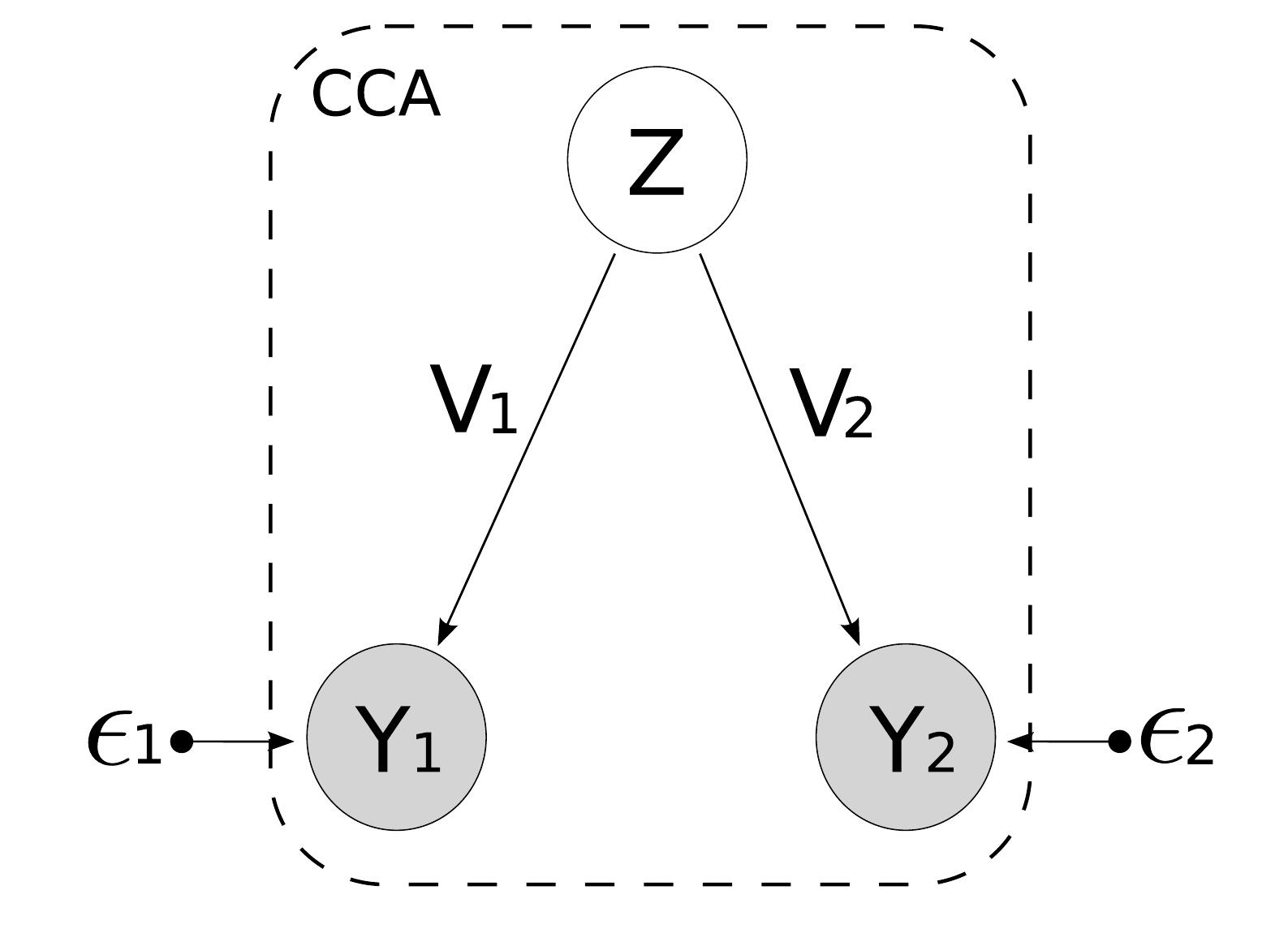}
}
\subfigure[]{ \label{fig:rcacca}
  \includegraphics[width=0.29\linewidth]{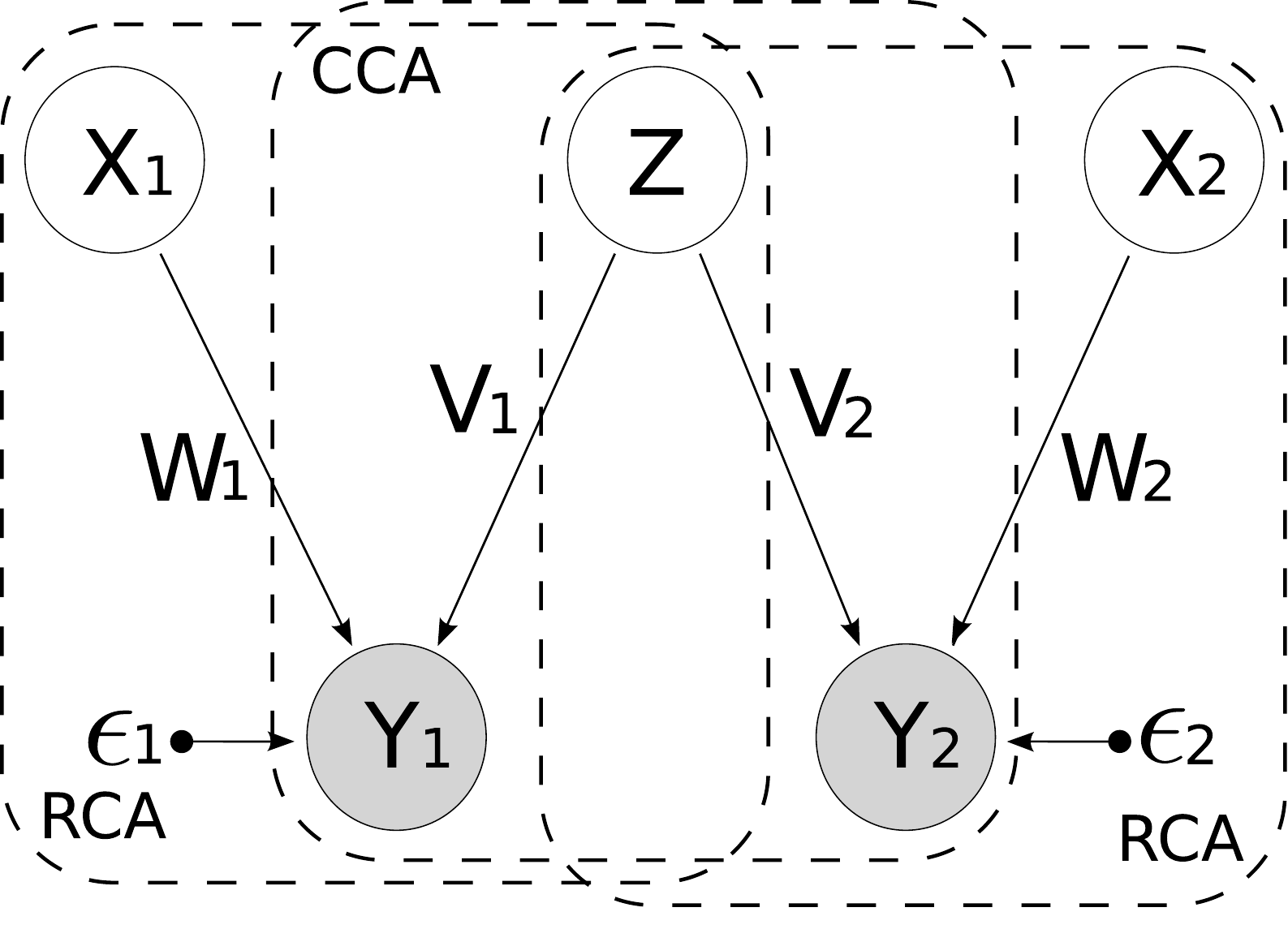}
}
\caption{
\small{
\subref{fig:rca}
Graphical model of RCA. $\mathbf{Z}$ partially explains the variation in
$\mathbf{Y}$ through the sub-space defined in $\mathbf{V}$. The \emph{residual}
covariance is spanned by $\mathbf{W}$ up to some noise variance of
$\mathbf{\epsilon}$.
\subref{fig:cca}
Graphical model of PCCA. $\mathbf{Z}$ is latent and shared between
$\mathbf{Y}_1$ and $\mathbf{Y}_2$. The standard linear approach to estimating
$\mathbf{V}_1, \mathbf{V}_2$ and $\mathbf{Z}$ is via CCA, which turns out to be
a special case of RCA (cf. section \ref{subsec:CCAequiv}).
\subref{fig:rcacca}
In a multi-view learning context, $\mathbf{Y}_1, \mathbf{Y}_2$ have shared
($\mathbf{Z}$) and private ($\mathbf{X}_1$, $\mathbf{X}_2$) latent components.
We illustrate an iterative-RCA approach for inference in this type of model.
}}
\label{fig:graphicalmodels}
\end{figure}

To show the equivalence of RCA to CCA, we turn our attention to the
generalised eigenvalue problem of RCA in eq. (\ref{eq:rcaeigenproblem_primal})
and consider the case where
\[
\mathbf{\Sigma}=\begin{pmatrix} \mathbf{Y}_1^\top\mathbf{Y}_1^{} & \mathbf{0} \\ 
\mathbf{0} & \mathbf{Y}_2^\top\mathbf{Y}_2^{}\end{pmatrix}.
\]
Then by inspection, the generalised eigenvectors $\mathbf{S}$ of RCA become the
\emph{canonical directions} and $(\mathbf{D-I})$ becomes the diagonal matrix of
\emph{canonical correlations}. \citep{Bach:pcca05} showed that the CCA maximum
likelihood solutions for $\mathbf{V}_1,\mathbf{V}_2$ in the graphical model of
Figure \ref{fig:cca} (again, for centred $\mathbf{y}_1,\mathbf{y}_2$)
\[
  \begin{pmatrix} \mathbf{y}_1 \\
  \mathbf{y}_2 \end{pmatrix} \sim
  \mathcal{N} \left(\begin{pmatrix} \mathbf{V}_1 \\
      \mathbf{V}_2  \end{pmatrix} \mathbf{z}, 
      \begin{pmatrix} \boldsymbol{\epsilon}_1 & \mathbf{0} \\
      \mathbf{0} & \boldsymbol{\epsilon}_2 \end{pmatrix} \right),
\]
are $\mathbf{V}_1 =
\mathbf{C}_{11}\mathbf{S}_{1}^{(q)}\mathbf{\Lambda}^{1/2}_q\mathbf{R}$ and
$\mathbf{V}_2 =
\mathbf{C}_{22}\mathbf{S}_{2}^{(q)}\mathbf{\Lambda}^{1/2}_q\mathbf{R}$, where
$\boldsymbol{\epsilon}_1$ and $\boldsymbol{\epsilon}_2$ are full noise
covariance matrices, $\mathbf{S}_1^{(q)}$ and $\mathbf{S}_2^{(q)}$ are the first
$q$ canonical directions, $\boldsymbol{\Lambda}_q$ is the diagonal matrix of the
first $q$ canonical correlations and the arbitrary rotations
$\mathbf{R}=\mathbf{I}$. These maximum likelihood solutions are equivalent to
$\mathbf{W}$ in eq. (\ref{eq:rcaeigenproblem_primal}), when
$\mathbf{\Sigma}=\mathbf{C}_{11}$, $\mathbf{S}=\mathbf{S}_1^{(q)}$ for
$\mathbf{V}_1$ and $\mathbf{\Sigma}=\mathbf{C}_{22}$,
$\mathbf{S}=\mathbf{S}_2^{(q)}$ for $\mathbf{V}_2$.

Similarly, we get the PCA eigenvalue equation when $\mathbf{\Sigma} =
\mathbf{I}$ and the PPCA solution emerges as
\begin{equation*}
\mathbf{W} =  \mathbf{S}\mathbf{(D - I)}^{\frac{1}{2}}
= \mathbf{U}_q(\mathbf{\Lambda}_q - \mathbf{I})^{\frac{1}{2}}.
\end{equation*}
We notice a subtle difference from the PPCA formulation here. Whereas PPCA
explicitly subtracts the noise variance from the $q$ retained principal
eigenvalues, RCA already incorporates any noise terms in $\mathbf{\Sigma}$ and
\emph{standardises} them while projecting the total covariance onto the
eigenbasis of $\mathbf{\Sigma}$, see eq. (\ref{eq:projectedK}).

From the RCA perspective, CCA can be seen as setting $\mathbf{\Sigma}$ to be
block diagonal, with each block containing the sample covariance
matrix associated with the data. The residual components then
represent the variance which isn't explained by those two sample
covariances: i.e. the correlation between the two data sets. Residual
components analysis is much general than this though, by alternative
choices for $\mathbf{\Sigma}$ we can explore other residual components. To
demonstrate this we now consider two case study data sets. The first
is a gene expression experiment containing treatment and control, our
objective will be to explore the differences between treatment and
control. The second is a data set of human pose and silhouette
\citep{Agarwal:pose06}. Our objective is to predict the pose given the
silhouette and we find a set of components which we can project the data on to
achieve this.


\section{Case Study 1: Differences in Gene Expression Profiles}
\label{subsec:geneexpression}

A common data analysis challenge is to summarize the difference
between treatment and control. To illustrate how RCA can help, we
consider two gene expression time series of cell lines. The treatment
cells are targeted by TP63 introduced into the nucleus by
tamoxifen. The control cells are simply subject to tamoxifen
alone. The data used for this case study come from
\citep{della2008direct}\footnote{Data is available on the Gene
  Expression Omnibus (GEO) database, under accession number
  GSE10562.}.  The treatment group ($\mathbf{Y}_1$) contains $n_1=13$
time points of $d=22,690$ gene expression measurements, whilst the
control group ($\mathbf{Y}_2$) contains only $n_2=7$ time points. This
complexity of data (with different numbers of time points and
non-uniform sampling) is typical of many bio-medical data sets. The
challenge is to represent the differences between the gene expression
profiles for these two data sets. Canonical correlates analysis could
be applied but this would represent the similarities between the data
not the differences. Our approach is as follows. First we assume that
both time series are identical, that would imply that they could be
modeled (for example) by a Gaussian process with a temporal covariance
function,
\[ 
\begin{pmatrix}\mathbf{y}_1\\ \mathbf{y}_2\end{pmatrix} \sim
\mathcal{N}(\mathbf{0}, \mathbf{K})
\]
where the matrix of the covariance function, $\mathbf{K}$, is computed
as if both $\mathbf{y}_1$ and $\mathbf{y}_2$ were from the same
function. Now, if we study the residual components, they will be forced
to explain how the two time series are actually different. In other
words we model the data through the \emph{dual} paradigm with a
covariance of the form
\[
\tilde{\mathbf{C}} = \mathbf{X}\mathbf{X}^\top + \mathbf{K}
\] 
and solve to find the residual components $\mathbf{X}$.
We used a squared exponential covariance (or RBF kernel) for
$\mathbf{K}$ whose elements were $k(t_i, t_j) = \exp(-0.5\ell^{-2}(t_i
- t_j)^2)$. The parameters of the covariance function could be optimized, but for
simplicity we set $\ell=20$ which provided a bandwidth roughly in line with the
time point sampling intervals. We also added a small noise term along the
diagonal of $\mathbf{K}$ which was set to 1\% of the data variance.

We project the profiles onto the eigenbasis of the first $q$ generalised
eigenvectors
\[
 \mathbf{Y}^\prime = \mathbf{S}^{(q)\top}\mathbf{Y}
\]

and obtain a score of differential expression based on the norms of
their projections. The number $q$ of retained principal eigenvectors is decided
on the number of corresponding eigenvalues larger than one. Recall in
PPCA (cf. page \pageref{sec:intro}) that as the assumed noise variance 
$\sigma^2_n$ increases, more eigenvalues become negative and less eigenvectors
are retained in the solution of $\mathbf{W}$. On a similar note, RCA
standardises any positive-definite noise (cf. eq. (\ref{eq:projectedK})), so we
always have to test for eigenvalues larger than 1. Here, the assumed noise
variance embedded in the kernel drives the effective number of eigenvectors in
the projection.

\begin{figure}[tb]
\centering
\subfigure[]{ \label{fig:combRBF}
  \includegraphics[height=0.3\linewidth]{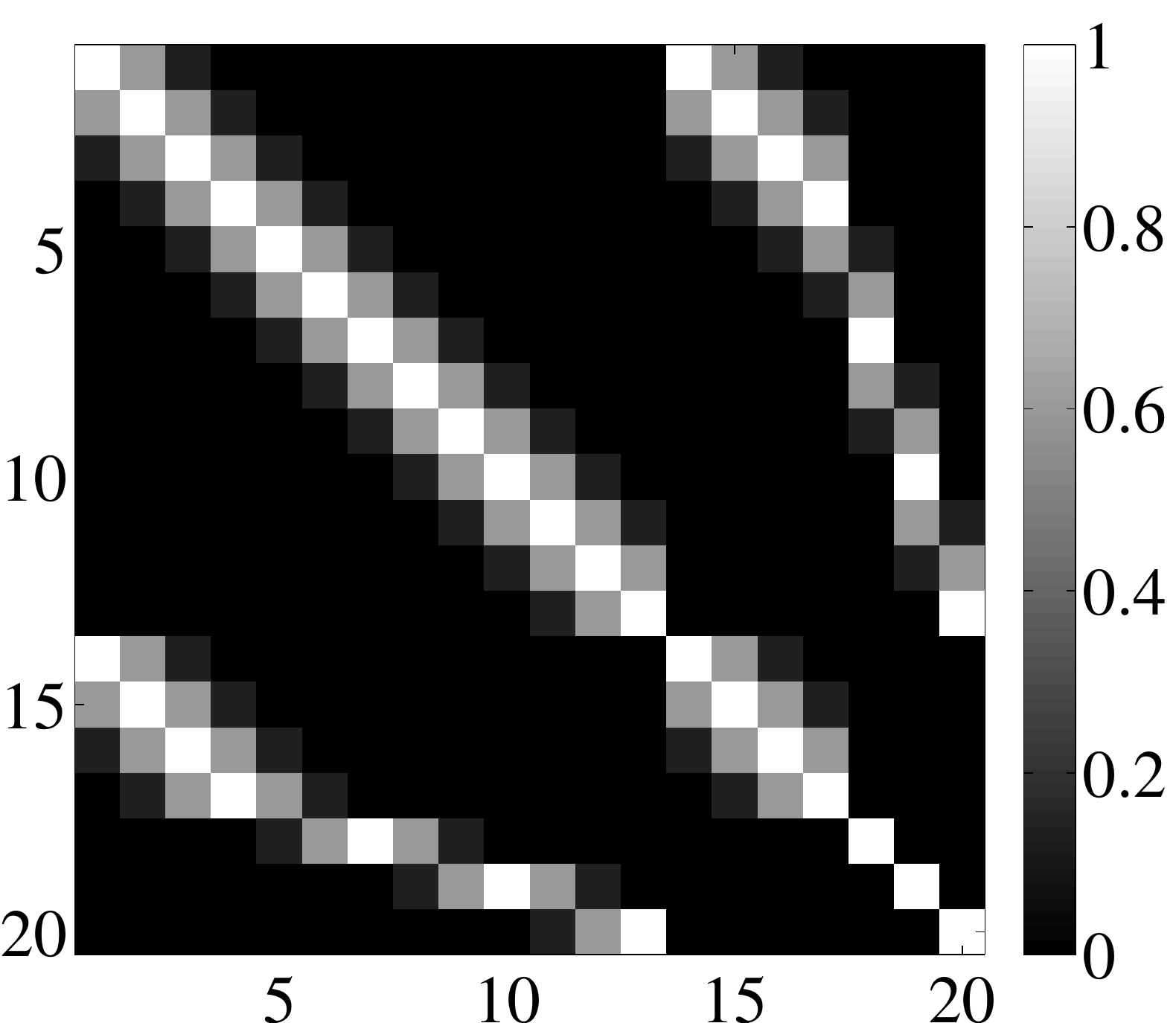}
}
\subfigure[]{ \label{fig:RCAvsBATS}
  \includegraphics[height=0.3\linewidth]{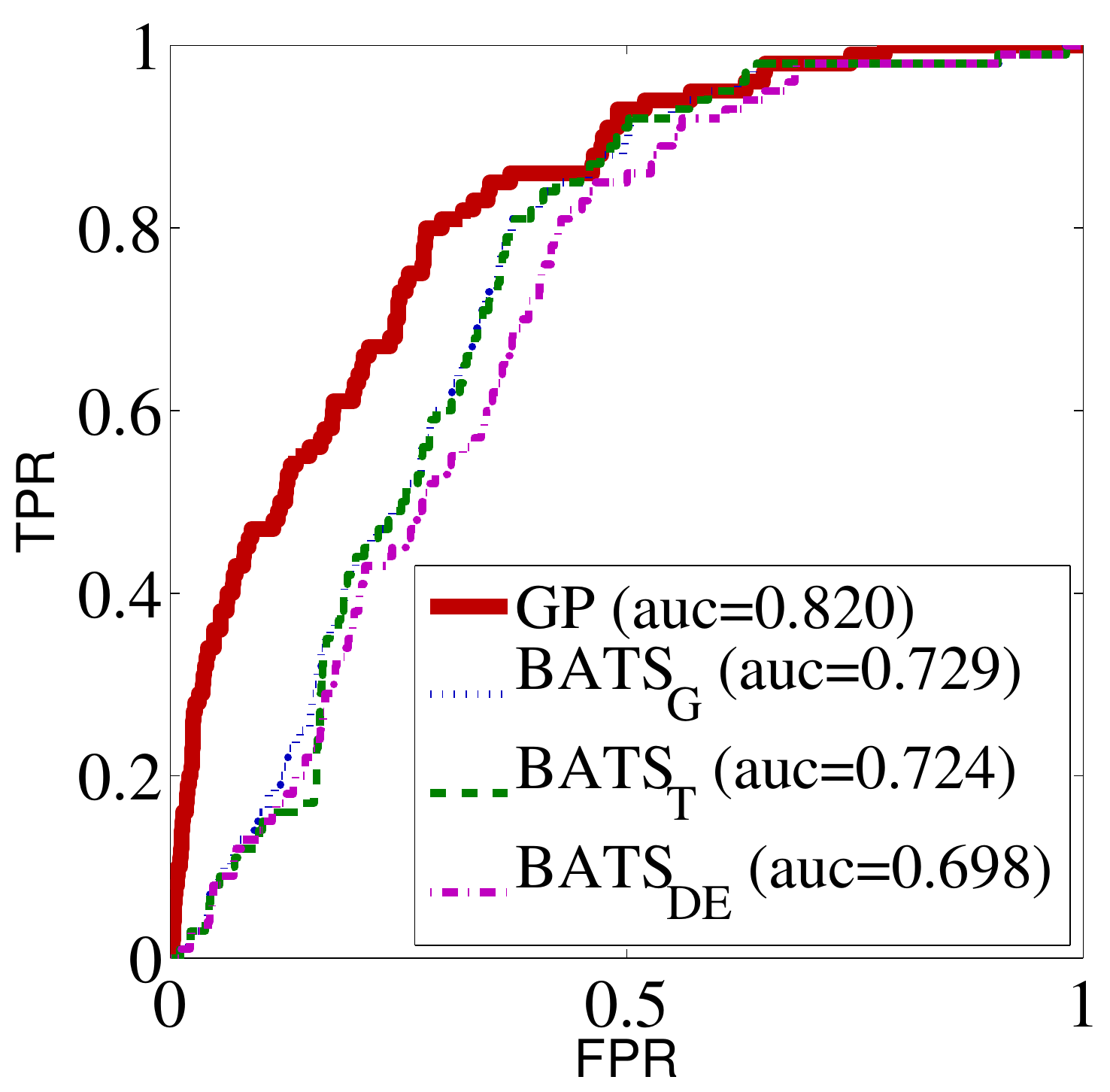}
}
\caption{ \subref{fig:combRBF} An RBF covariance computed on the
  augmented time-input vector for the microarray experiment. The
  covariance is computed across the times for the control and the
  treatment.  $\mathbf{t} = ({0:20:240}, 0, 20, 40, 60, 120, 180,
  240)$, with bandwidth parameter $\ell = 20$ and noise variance
  $\sigma^2_n = 10^{-4}$.  \subref{fig:RCAvsBATS} ROC comparison
  against BATS, see also \cite{Kalaitzis:simple11} for details on an
  alternative approach based on Gaussian processes.}
\end{figure}

We rank the scores and compare to a noisy ground truth list of binding targets
of TP63\footnote{A gene with a high number of binding sites for TP63 is a strong
candidate for being one of its direct targets (i.e. associated with TP63 related
diseases). The ranking list of direct targets is available at \url{
genome.cshlp.org/content/suppl/2008/05/05/gr.073601.107.DC1/DellaGatta_SupTable1
.xls}}
from \citep{della2008direct}, giving the ROC performance curve in
Figure \ref{fig:RCAvsBATS}. The baseline method that we compare against is a
Bayesian hierarchical model, BATS\footnote{The software of Bayesian Analysis for
Time Series is available at
\url{http://www.na.iac.cnr.it/bats/index_file/download.htm}.}
\citep{angelini2007bayesian}. We notice that RCA outperforms BATS in
terms the area under the ROC curve.

\section{Case Study 2: Iterative RCA for Prediction of Pose from Silhouette}
\label{sec:pose}

Probabilistic canonical correlates analysis explains two related data
sets by assuming a full covariance block diagonal form and low rank
off diagonal terms. Ek \emph{et al.}\cite{Ek:ambiguity08} introduced a model
with both a shared latent space and private latent spaces for explaining data
specifically associated with the two data sets. The graphical model is
shown in Figure \ref{fig:rcacca}. Each partition of the data space,
$\mathbf{Y}_1$ and $\mathbf{Y}_2$ has its own associated latent space,
$\mathbf{X}_1$ and $\mathbf{X}_2$ as well as a shared latent space,
$\mathbf{Z}$ which corresponds to the standard shared latent space
found in CCA. The advantage to a model of this structure is that
if the variance that is particular to each partition of the data is
low dimensional, this will be recovered. The partitions of the data
are therefore modeled as
\[
\mathbf{Y}_1 = \mathbf{X}_1\mathbf{W}_1^\top + \mathbf{Z}\mathbf{V}_1^\top +
\epsilon_1, \quad \textrm{where} \quad \epsilon_1
\sim \mathcal{N}(\mathbf{0},\sigma_1^2 \mathbf{I})
\]
and 
\[
\mathbf{Y}_2 = \mathbf{X}_2 \mathbf{W}_2^\top + \mathbf{Z} \mathbf{V}_2^\top +
\epsilon_2 \quad \textrm{where} \quad \epsilon_2
\sim \mathcal{N}(\mathbf{0}, \sigma_2^2 \mathbf{I}). 
\]
Each set of latent variables can be marginalized through an isotropic
Gaussian prior, $z \sim \mathcal{N}(0, 1)$, leading to a covariance
structure for the concatenated data set of the following form
\[
\tilde{\mathbf{C}} = \begin{pmatrix} \mathbf{W}_1^{}\mathbf{W}_1^\top & \mathbf{0}\\
\mathbf{0} & \mathbf{W}_2^{} \mathbf{W}_2^\top \end{pmatrix} + \begin{pmatrix}
\mathbf{V}_1^{}\mathbf{V}_1^\top & \mathbf{V}_1^{}\mathbf{V}_2^\top \\
\mathbf{V}_2^{}\mathbf{V}_1^\top & \mathbf{V}_2^{} \mathbf{V}_2^\top
\end{pmatrix} +
\begin{pmatrix}
\sigma^2_1 \mathbf{I} & \mathbf{0} \\
\mathbf{0} & \sigma^2_2 \mathbf{I}
\end{pmatrix}.
\]
Setting 
\[\mathbf{\Sigma} = \begin{pmatrix} \mathbf{W}_1\mathbf{W}_1^\top &
  \mathbf{0}\\ \mathbf{0} & \mathbf{W}_2{} \mathbf{W}_2^\top \end{pmatrix} +
\begin{pmatrix}
\sigma^2_1 \mathbf{I} & \mathbf{0} \\
\mathbf{0} & \sigma^2_2 \mathbf{I}
\end{pmatrix}
\]
allows
$\mathbf{V}_1$ and $\mathbf{V}_2$ to be optimized using the RCA
algorithm. To optimize $\mathbf{W}_1$ and $\mathbf{W}_2$ we note that
the marginal covariance for $\mathbf{Y}_1$ is $ \tilde{\mathbf{C}}_{11} =
\mathbf{W}_1\mathbf{W}_1^\top + \mathbf{V}_1\mathbf{V}_1^\top +
\sigma^2_1 \mathbf{I} $, so $\mathbf{W}_1$ can be optimized by RCA using
$\mathbf{\Sigma}= \mathbf{V}_1\mathbf{V}_1^\top + \sigma^2_1 \mathbf{I}$. A
similar optimization can be done for $\mathbf{W}_2$.

The data we consider come from Agarwal and Triggs
\citep{Agarwal:pose06}. They produced a set of 3D human poses and
associated silhouettes. The silhouettes are summarized by a $d_2=100$
dimensional vector of HoG features in matrix $\mathbf{Y}_2 \in \Re^{n
  \times d_2}$. There are $n = 1,927$ frames. There are 21 points in
each pose representation each containing $x$, $y$, $z$ coordinates
leading to $d_2=63$ for $\mathbf{Y}_1 \in \Re^{n \times d_1}$.  The
data is generated by the Poser computer software, therefore it is
``noise free''. To better reflect real world scenarios we added a
small amount of Gaussian noise to each feature.

One issue with this iterative RCA algorithm is that 3 latent
dimensionalities need to be chosen. However, similar to probabilistic
PCA, if the noise values, $\sigma_1^2$ and $\sigma_2^2$ are fixed, the
latent dimensionality will be determined automatically.  We therefore
set the noise variances to a proportion, $\alpha$, of the data
variance. We used this fraction to control the dimensionality, varying
it between 0 and 1. This gave us only one parameter in the model to
vary. The algorithm converges when the log-marginal likelihood between
two iterations differs no more than a small constant.
\begin{algorithm}
\caption{Iterative RCA}
\label{iterativeRCA}
\begin{algorithmic}
\STATE $\mathbf{C} \leftarrow n^{-1}\mathbf{Y}^\top\mathbf{Y}, \quad
\mathbf{C}_{11} \leftarrow n^{-1}\mathbf{Y}_1^\top\mathbf{Y}_1^{},  \quad
\mathbf{C}_{22} \leftarrow n^{-1}\mathbf{Y}_2^\top\mathbf{Y}_2^{}$
\STATE Initialize $\alpha \in [0,1], \quad \sigma^2_1 \leftarrow
\frac{\alpha}{d_1} \ \textrm{tr}(\mathbf{C}_{11}),\  \sigma^2_2 \leftarrow
\frac{\alpha}{d_2} \ \textrm{tr}(\mathbf{C}_{22}), \quad 
(\mathbf{W}_1, \mathbf{W}_2 , \mathbf{V}_1, \mathbf{V}_2) \leftarrow \
\mathbf{0}$
\REPEAT
\STATE Compute \quad $\tilde{\mathbf{W}}_1$ \quad by \quad
$ \mathbf{C}_{11} \mathbf{\tilde{W}}_1 = (\mathbf{V}_1^{}\mathbf{V}_1^\top +
\sigma^2_1\mathbf{I})\mathbf{\tilde{W}}_1\mathbf{\Lambda}_1$
\STATE $\mathbf{W}_1 \quad \leftarrow \quad (\mathbf{V}_1^{}\mathbf{V}_1^\top +
\sigma^2_1\mathbf{I})
\mathbf{\tilde{W}}_{1}^{(q)} (\mathbf{\Lambda}_{1}^{(q)} -
\mathbf{I})^{\frac{1}{2}}$
\STATE Compute \quad $\mathbf{\tilde{W}}_2$ \quad by \quad
$ \mathbf{C}_{22} \mathbf{\tilde{W}}_2 = (\mathbf{V}_2^{}\mathbf{V}_2^\top +
\sigma^2_2\mathbf{I})\mathbf{\tilde{W}}_2\mathbf{\Lambda}_2$
\STATE$ \mathbf{W}_2 \quad \leftarrow \quad (\mathbf{V}_2^{}\mathbf{V}_2^\top +
\sigma^2_2\mathbf{I})
\mathbf{\tilde{W}}_{2}^{(q)} (\mathbf{\Lambda}_{2}^{(q)} -
\mathbf{I})^{\frac{1}{2}}$
\STATE $\mathbf{\Sigma} \quad \leftarrow \quad  \begin{pmatrix}
\mathbf{W}_1^{}\mathbf{W}_1^\top + \sigma^2_1\mathbf{I} & \mathbf{0} \\
\mathbf{0} & \mathbf{W}_2^{}\mathbf{W}_2^\top + \sigma^2_2\mathbf{I}
\end{pmatrix}$, \quad compute \quad $\tilde{\mathbf{V}}$ \quad by \quad
$\mathbf{C} \tilde{\mathbf{V}} =  \mathbf{\Sigma}\mathbf{\tilde{V}}
\mathbf{\Lambda}$
\STATE $\mathbf{V} \quad \leftarrow \quad \mathbf{\Sigma} \mathbf{\tilde{V}}^{(q)}
(\mathbf{\Lambda}^{(q)} - \mathbf{I})^{\frac{1}{2}},
\quad \mathbf{V}_1 \quad \leftarrow \quad \mathbf{V}_{1:d_1,:},
\quad \mathbf{V}_2 \quad \leftarrow \quad \mathbf{V}_{(d_1+1):(d_1+d_2),:}$
\UNTIL{the log-marginal likelihood converges}
\end{algorithmic}
\end{algorithm}
The prediction of pose from silhouette can be computed through
$p(\mathbf{Y_1}|\mathbf{Y}_2)$. The mean of this density is given by
\[
 \mathbf{y}_1^* = \mathbf{V}_1 \mathbf{V}_2^\top (\mathbf{W}_2 \mathbf{W}_2^\top
+ \sigma^2_2)^{-1} \mathbf{y}_2  + \boldsymbol{\mu}_1,
\]
where $\boldsymbol{\mu}_1$ is the sample mean of $\mathbf{Y}_1$. Variances can
also be computed, but aren't used in our experiments.

\begin{figure}[tb]
\centering
\subfigure[]{ \label{fig:RCAvsPCCA}
  \includegraphics[height=0.3\linewidth]{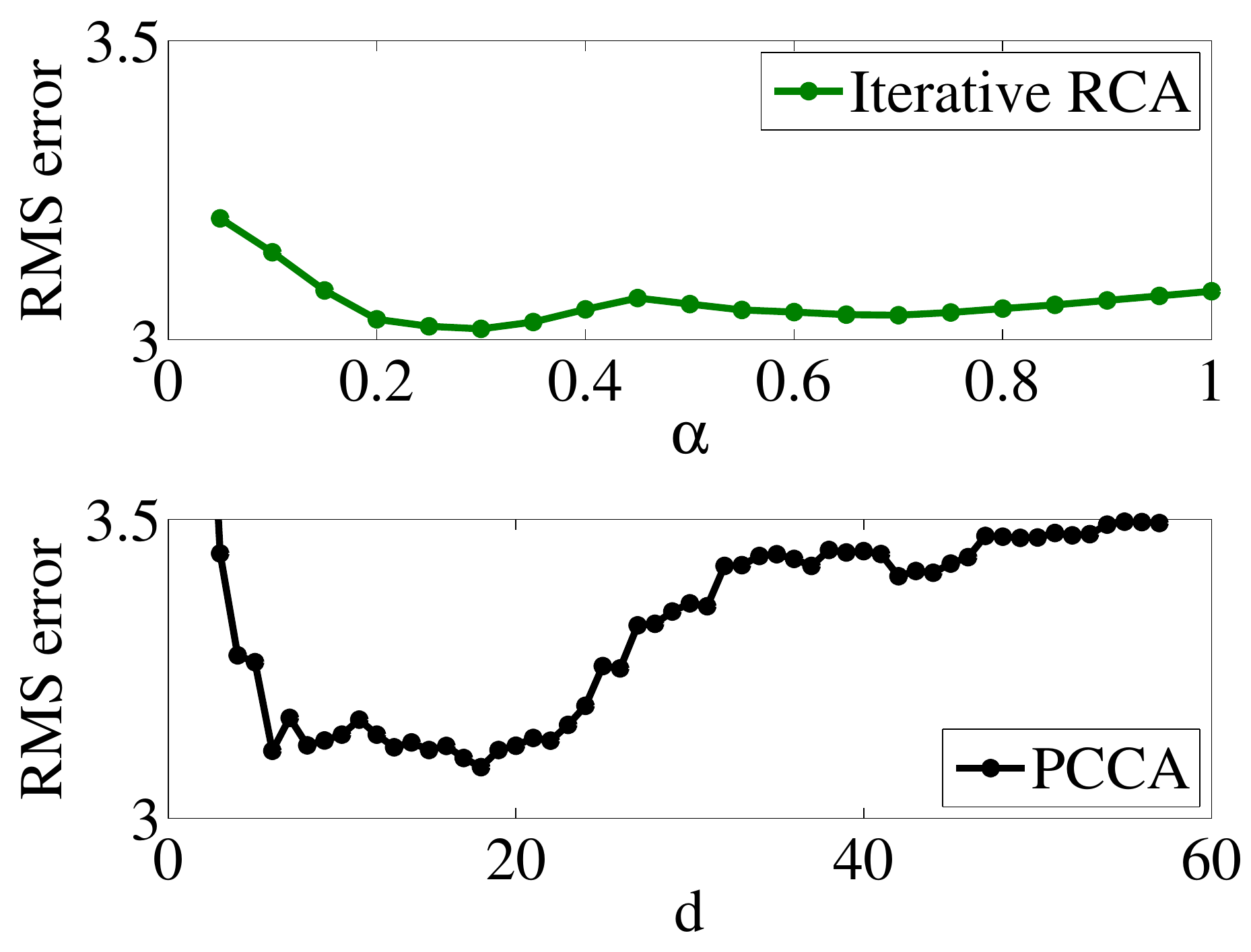}}
\subfigure[]{ \label{fig:RCAretdim}
  \includegraphics[height=0.3\linewidth]{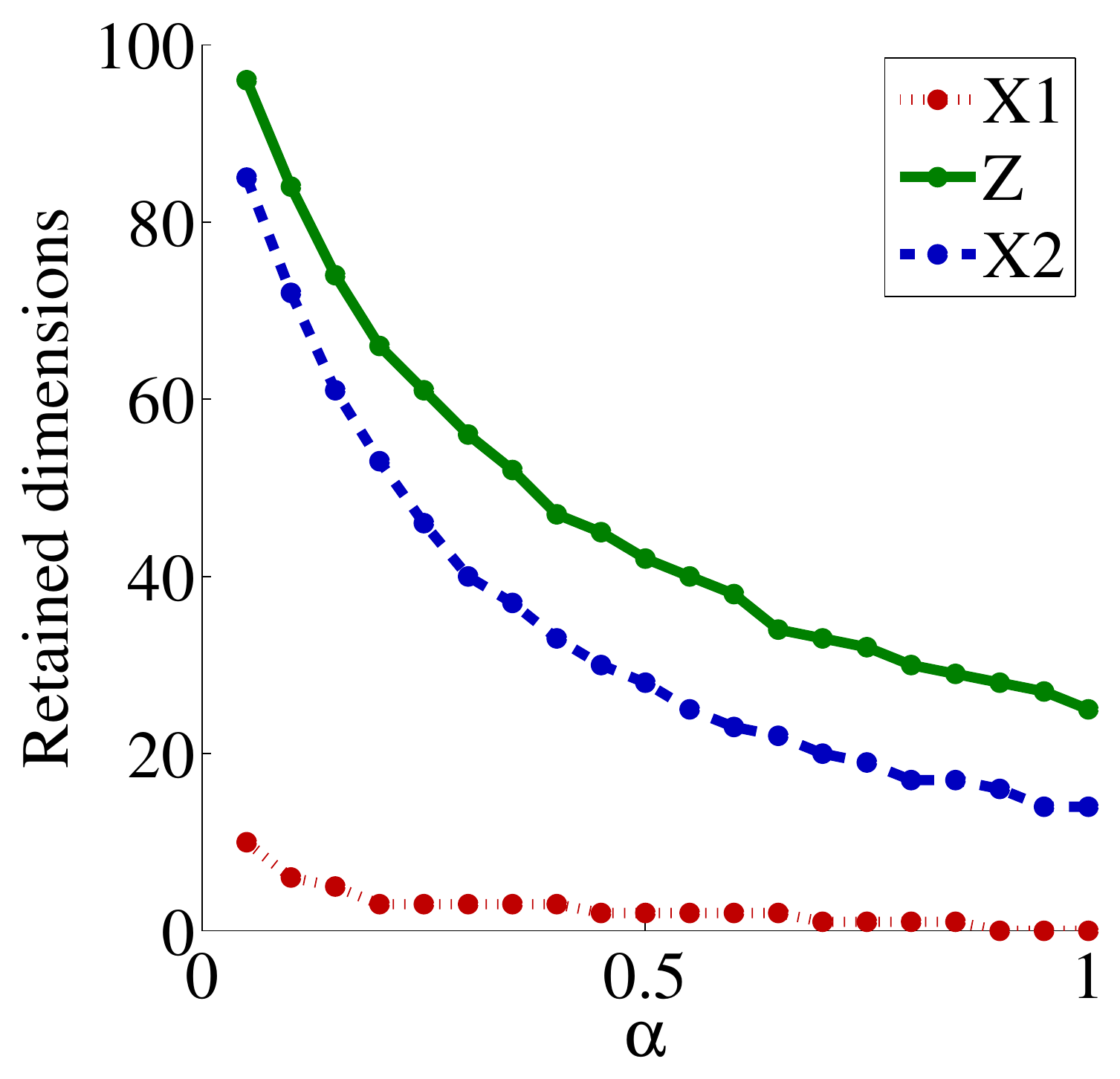}}
\vspace{0.5cm}
\subfigure[]{ \label{fig:skeletons}
  \includegraphics[width=0.9\linewidth]{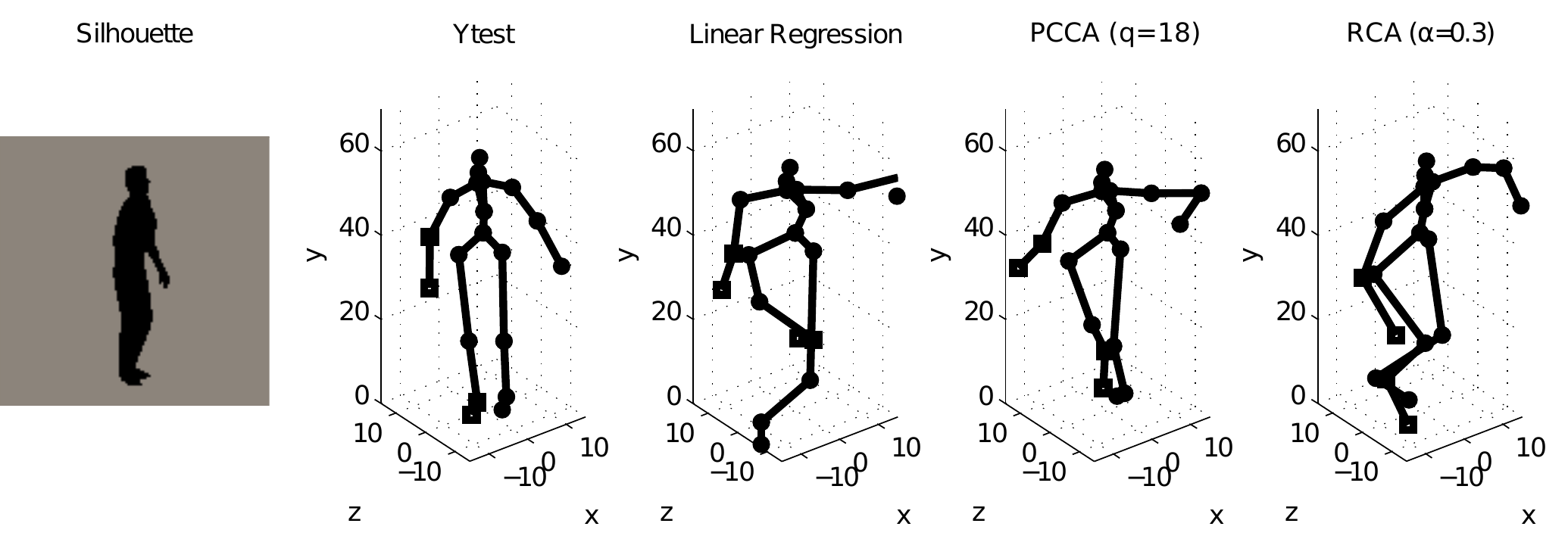}}

\caption{Comparison of iterative RCA for a shared latent space with standard CCA
and a linear regression model.
\subref{fig:RCAvsPCCA} Iterative RCA against standard probabilistic CCA with
root mean square errors for reconstruction of the pose. The figure shows the
error for varying $q$ (i.e. the latent dimensionality) in PCCA and varying
$\alpha$ (i.e. the proportion of explained variance) in  RCA. Linear regression
also yields an RMS = 3.2098. \subref{fig:RCAretdim} Latent dimensionalities on
convergence, of $\mathbf{X}_1$, $\mathbf{X}_2$ and $\mathbf{Z}$ (cf. Figure
\ref{fig:rcacca}), for varying $\alpha$.
\subref{fig:skeletons} Shows the silhouette and the pose predictions, alongside
ground truth for frame \#404, which was the frame in the test set with the
largest error.}
\end{figure}

Comparison of Iterative RCA with varying $\alpha$, to PCCA with
varying $q$, yields the root mean square (RMS) errors illustrated in
Figure \ref{fig:RCAvsPCCA}. Iterative RCA outperforms standard PCCA in
general with the smallest difference in performance being at $q = 18$
for PCCA and $\alpha = 0.3$. The RMS error of RCA is robust for a wide
range of large $\alpha$ values. An interesting aspect of iterative RCA
is the self-regularity that the algorithm imposes on the latent
dimensionalities of the shared and private components, see Figure
\ref{fig:RCAretdim}. As the noise increases with $\alpha$, the
eigenvalues decay faster from $\mathbf{Z}$ and $\mathbf{X}_2$ than
from $\mathbf{X}_1$. Other approaches to selecting the dimensionality of the
latent spaces could also be followed, but the approach of explaining a
proportion of the variance with the noise seems simple and satisfactory.

\section{Discussion}

We have introduced residual component analysis: an algorithm for
describing a low dimensional representation of the residuals of a data
set given partial explanation by a covariance matrix $\Sigma$. With
imaginative application our algorithm allows for novel approaches to
data analysis. We illustrated this with the characterization of the
difference between a treatment and control time series and an
algorithm for fitting a low dimensional variant of CCA. Other forms of
$\mathbf{\Sigma}$ that could be of interest include one with a sparse inverse.
Sparse inverse structures capture relations between variables that are not well
characterized by low rank forms. As such, the combination of sparse inverse and
low rank could be a powerful one. Finally a form which reflects class structure
in the data would also allow the exploration of components of the data which
were not related to the class structure.

\footnotesize{\small}
\bibliographystyle{plainnat}
\renewcommand*{\refname}{\begin{normalsize}References\end{normalsize}}
\bibliography{rcaArXiv-MATweave}


\end{document}